\pdfoutput=1 
\documentclass{article}

     \usepackage[preprint]{metalearning_skin_cancer_dl}

 \usepackage[]{metalearning_skin_cancer_dl}

\usepackage[utf8]{inputenc} 
\usepackage[T1]{fontenc}    
\usepackage{hyperref}       
\usepackage{url}            
\usepackage{booktabs}       
\usepackage{amsfonts}       
\usepackage{nicefrac}       
\usepackage{microtype}      
\usepackage[ruled,vlined]{algorithm2e}
\include{pythonlisting}
\usepackage{graphicx}
\graphicspath{ {./graphs/} }

\title{Meta-learning for skin cancer detection using Deep Learning techniques}
\author{
 Sara I. Garcia\\
  Faculty of Engineering, Environment and Computing\\
  Coventry University\\
  Coventry, UK \\
  \texttt{garciaas@coventry.ac.uk} \\
}

\begin{document}

\maketitle

\begin{abstract}This study focuses on automatic skin cancer detection using a Meta-learning approach for dermoscopic images. The aim of this study is to explore the benefits of the generalization of the knowledge extracted from non-medical data in the classification performance of medical data and the impact of the distribution shift problem within limited data by using a simple class and distribution balancer algorithm. In this study, a small sample of a combined dataset from 3 different sources was used to fine-tune a ResNet model pre-trained on non-medical data. 
The results show an increase of performance on detecting melanoma, malignant (skin cancer) and benign moles with the prior knowledge obtained from images of everyday objects from ImageNet dataset by 20 points. These findings suggest that features from non medical images can be used towards the classification of skin moles and that the distribution of the data affects the performance of the model.
\end{abstract}

\section{Introduction}

Deep learning is a technique that has been widely used in the fields of image classification and object detection due to its ability to learn highly complex patterns from vast amounts of data with reliable performance. One of the greatest challenges in the application of deep learning in medical domains is the limited availability of labelled data. This is mainly because it represents a considerable cost and time for a professional clinician to assess the medical condition of the patient.One of the techniques used to deal with this limitation is data augmentation, which consists in generating artificial examples from the target data, although this does not overcome the problem \citep{Wong2016, Mikolajczyk2018}. Another technique which has been widely used is Transfer learning. In this approach, the network is first trained with data from a related domain, the network weights are then transferred to a new network and fine-tuned with the target data. The drawback of this method is that it presents poor performance when the amount of the target data is too small or presents a slight distribution shift in the target data \citep{Soekhoe2016, Kumar2019}.
Despite that meta-learning is not a new concept in the deep learning community, it has not been widely explored for the application in the medical imaging domain. In this study, an exploration of a meta-learning model is reviewed, emphasizing in the dataset bias and distribution shift, factors that have been extensively addressed in image recognition problems, but have not received a significant importance for applications on medical imaging data.

\subsection{Problem overview}

Melanoma is the fifth most common cancer in the UK. Since 1990, its incidence has increased 134\% in the UK. It is estimated that around 2,400 deaths are caused by melanoma each year in the UK. However, the survival rate is 90\% when diagnosed on time. 

This suppose a relevant problem that can be solved with Deep Learning. The main challenge in the application of Deep Learning in the medical domain is the scarcity of the data. This limitation can be addressed with meta-learning approaches that generalize the knowledge of unrelated domains to help on the prediction of the task.
Since the recent advancements in the development of meta-learning frameworks, meta-learning has started to attract researchers from the medical imaging community. A first exploration of the viability of using meta-learning approaches in medical domains is explored in this study.

\subsection{Meta-learning concepts and definitions}

The term Meta-learning, also known as "Learning to Learn” has its origins in educational psychology, where it is described as the adaptation of the learning process according to the requirements of a specific task \citep{BIGGS1985, Lemke2013}.
The goal of Meta-learning is to understand the learning process and exploit the acquired knowledge to improve the effectiveness of learning new tasks. In the context of artificial intelligence systems, a task can be a regression, object detection or classification problem, among all.
The knowledge derived from the learning mechanism is called Meta-knowledge, and the knowledge about which attributes are relevant to perform a task is called Metadata. A Meta-learning system uses the acquired meta-knowledge to derive the appropriate strategy to learn on a new domain of application, in other words, meta-learning is the type of learning that uses prior experience of other tasks to adapt to learn the new.

\section{Related Work}
\label{headings}

The increasing popularity of deep learning in object recognition tasks is primarily due to the availability of training data. However, in medical domains, the researchers are facing with data scarcity, which have put a research direction in using data from non medical domains to diagnose diseases from medical imaging data.\\

 \citep{DBLP:journals/corr/CheplyginaMVBP17} evaluated whether a classifier is able to predict which classification problem a dataset is sampled from, based on the performance of 6 different classifiers and 120 datasets from 6 different classification problems. Their findings show that different datasets from the same task share similar properties, such as dataset size, type of images, number of classes and type of features.
Despite the limitations of their approach, this simple method was able to extract features that are shared among datasets and find clusters. This demonstrates that there must be some intrinsic characteristics not only at the meta-level, but also between the samples of the datasets of similar tasks. \\

\citep{Schlegl2014} developed a convolutional neural network to classify pathologies in high-resolution Computed Tomography (CT) scans of lung tissue with partial annotations obtained from different sources. The authors report an improvement in the classification performance with the model pre-trained with natural images compared to the pre-training on medical data of a different task. Their results prove that the data from similar domains do not necessarily lead to an improvement in the learning, but instead, is the variability of the input data, such as colours, textures, shapes and angles, that led to a better generalization.\\

\citep{HU} implemented Reptile, a state-of-the-art meta-learning model pre-trained with mini-ImageNet to detect diabetic retinopathy. Their results show a slightly increment in the performance of the meta-learning model on detecting the target class as compared to the baseline model with no pre-training. This demonstrates that the amount of training data does not correlate with the classification performance. In other words, it is possible to obtain a good generalization from a small dataset size.\\

\citep{Esteva2017} implemented a GoogleNet Inception V3 model pre-trained on ImageNet. For this study, the authors designed a partitioning algorithm to balance the classes in the training set to avoid in-class bias. Their reports indicate that the transfer learning model is able to match the performance of the dermatologists on the critical diagnostic tasks. The performance of the model increases when the training data is balanced, which is an indicator of the presence of bias. This is studied in detail in \citep{Dietterich95machinelearning}, where the authors present a model for domain adaptation to overcome the covariate shift, by learning a representation of the data that takes into account the distribution shift between the test and training data. This problem was also addressed by \citep{DBLP:journals/corr/abs-1812-01716} who present a method to handle dataset bias for medical image data by unlearning the membership of the dataset samples using a \textit{leave-one-dataset-out} strategy. Both results shows a considerable improvement in the performance of an unbiased classification model.

In summary, the popularity of meta-learning in the application on medical domain data still need to be explored, but represents a promising research direction for the medical community.

\section{Dataset description}

The images used for this study correspond to three datasets from different sources. The size of the three datasets corresponds to 27531 dermoscopic images, from which only a sample of 193 images were used. The description of each dataset used in this study is given below:

\begin{itemize}
\item ISIC 2019: This is a public dataset for skin lesion analysis towards melanoma detection on high resolution dermoscopic images. For this study, only a subset from the training set was used.
\item PH2 Database: This dataset is composed by dermoscopic images acquired at the Dermatology Service of Hospital Pedro Hispano in Matosinhos, Portugal. This image database contains 200 dermoscopic images of 768x560 pixels.
\item 7-Point criteria evaluation database: This database is composed of a diagnosis and seven point checklist criteria labels publicly available from the Simon Fraser University website. This dataset is composed by over 2,000 dermoscopy images that correspond to twenty classes.
\end{itemize}
The images from the three datasets were chosen randomly, which corresponds to 1.2\% of the total size of the datasets. The labels of the three datasets were combined into three groups: melanoma, malignant and benign. The malignant class contains labels of skin cancer moles, such as basal cell carcinoma, squamous cell carcinoma and actinic keratosis. The rest of non-melanoma and non-cancer labels were grouped into the benign class. The class distribution of the combined dataset used for this study is given in Table~\ref{table1-dataset}.

\begin{table}
  \caption{Class distribution of the combined dataset}
  \label{table1-dataset}
  \centering
  \begin{tabular}{lll}
    \cmidrule(r){1-2}
    Class     & Number of Samples   
    \\
    \midrule
    \midrule
    Benign & 132 images    
    \\
    Malignant & 15 images    
    \\
    Melanoma & 46 images
    \\
    Total & 193 images
    \\
    \bottomrule
    \\
  \end{tabular}
\end{table}

\label{headings}

\section{Methodology}
\label{others}

For this study, a ResNet50 model was used for all the experiments. This is a variation of the ResNet architecture that consists of 50 convolutional layers. This architecture was used for its high performance on ImageNet and availability in the Keras library.
All the samples of the combined dataset were normalized and reduced by 254 x 254 pixels to match the input of ResNet. No image segmentation and colour variation were performed in this study. The parameters used in the experiment are given in Table~\ref{table2-parameters}.

\begin{table}
  \caption{Model parameters}
  \label{table2-parameters}
  \centering
  \begin{tabular}{lll}
    \midrule
    \midrule
    Learning rate & 0.0001    
    \\
    Loss & categorical cross entropy    
    \\
    Momentum & 0.9
    \\
    Optimizer & Stochastic Gradient Descent 
    \\
    \bottomrule
  \end{tabular}
\end{table}

Jaccard Similarity index and F1-score were used to measure the performance of the model on each experiment. Jaccard Similarity index was chosen for its wide use in the research community and therefore to serve as a point of comparison with the related works. F-1 score was chosen as it is a useful metric in the presence of class imbalance.

\section{Experiments and Results}
\label{others}

A set of three different experiments were conducted for this study. These experiments are listed below.
\begin{itemize}
    \item{ Bias detection}
    \item{ Bench-marking experiments on ResNet model with random weight initialization.}
    \item{ Pre-trained ResNet model fine-tuned with medical data.}
\end{itemize}

\subsection{Bias detection}

The \textit{Name that dataset!} experiment was conducted to detect the presence of dataset bias. This experiment was first designed by \citep{Khosla2012}. The experiment consists of measuring the performance of a classification model on recognizing which dataset an image comes from. If the model is able to recognize the dataset of origin, then we can say that the model is learning intrinsic characteristics particular to each dataset, or dataset bias.For this experiment, a simple convolutional neural network was implemented to detect the dataset bias. The results in table~\ref{table3-bias} show that the model is not able to infer the membership of the samples. This result is in alignment with our initial hypothesis, that no substantial difference between the samples of each dataset could be seen by the model, since the images of the three datasets were taken with the same device, and possibly under similar conditions.

\begin{table}
  \caption{Summarized results of bias detection experiments}
  \label{table3-bias}
  \centering
  \begin{tabular}{lll}
    \midrule
    \midrule
    Jaccard Similarity Index & 0.0    
    \\
    Overall precision & 0.32
    \\
    Overall recall & 0.32
    \\
    Overall F1-score & 0.32
    \\
    \bottomrule
    \\
  \end{tabular}
\end{table}

\subsection{Benchmarking experiments}

\begin{figure}
  \centering
  \fbox{
  \includegraphics[scale=0.3]{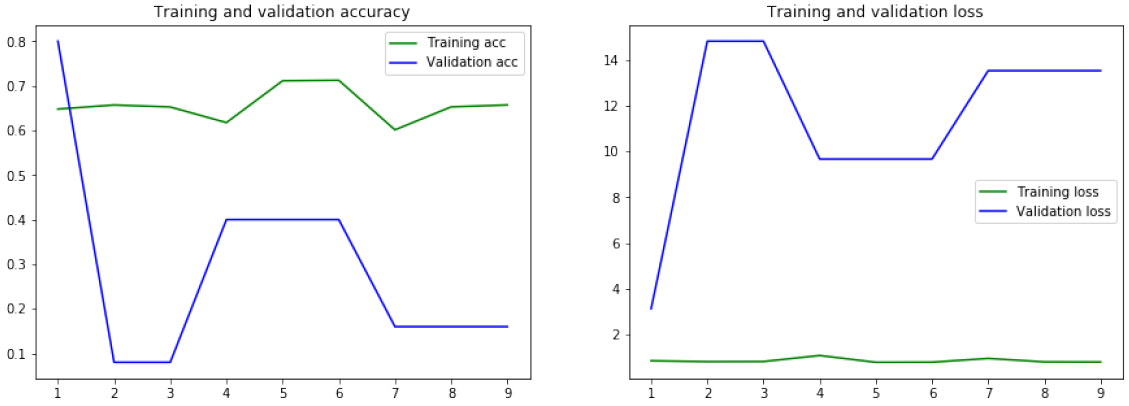}
   }
  \caption{Left image: training and validation loss of the benchmarking experiments. Right image: training and validation accuracy.}
  \label{fig1-benchmarking}
\end{figure}

For this experiment, a ResNet model was used with random weights initialization. This experiment was conducted to measure the performance of the baseline model with only the knowledge learned from the medical target data. The results of this experiment are given in Figure~\ref{fig1-benchmarking}. They were averaged out of 3 times using cross validation with stratified folds to ensure the class balance.

\subsection{Meta-learning experiments}

For this set of experiments, a pre-trained ResNet model was used. The weights of the network were obtained from training on the ImageNet dataset. The last softmax and dense layer was removed for the fine-tuning. The results of these experiments were averaged out of 3 times using cross validation with stratified folds. The following configurations were used for this set of experiments:
\begin{itemize}
    \item{Train and evaluation with no weighted classes.}
    \item{Train and evaluation with weighted classed.}
\end{itemize}

The combined dataset is highly imbalanced, which causes overfitting due to the limited amount of samples of the minority classes. This is also related with the distribution shift between the training and test datasets, which does not share the same distribution of the data. To alleviate this effect, a simple class and distribution balancer algorithm is proposed.
The purpose of this algorithm is to balance the classes in the training dataset and replicate the distribution of the test dataset in the training dataset to produce a set of weights for each class in the data that can be used during training. For its simplicity, it possess some drawbacks. One of them is that the same number of classes in the test set must exist in the training set and vice versa. See Algorithm~\ref{algorithm1}.

\begin{algorithm}[H]
\label{algorithm1}
\SetAlgoLined
Get majority class in Training set
\textbf{TR->MC}
 \\
 \ForEach C {
  Divide the number of samples of \textit{C} in \textit{TR} into the number of samples of \textit{C} in Test set \textbf{TE->DTC}
 
 Divide \textit{MC} into the numbers of samples of \textit{C} in \textbf{TR->ITC} 
 
 Set the weight for \textbf{C->DTC * ITC} 
 }
 
 \caption{simple class and distribution balancer algorithm}
\end{algorithm}

\begin{figure}
  \centering
  \fbox{
  \includegraphics[scale=0.5]{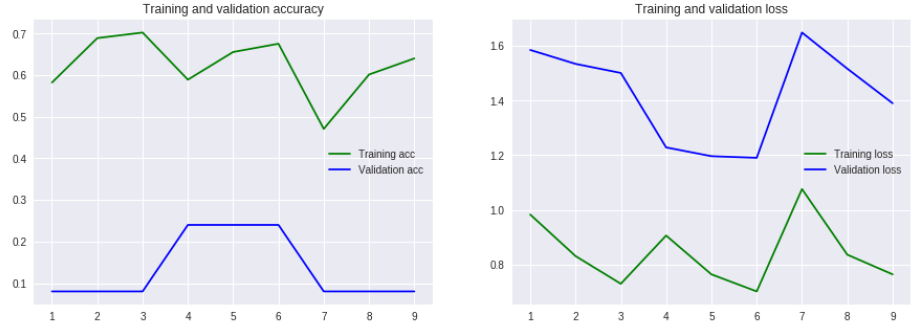}
  }
  \caption{to the right: training and validation accuracy. To the left: training and validation loss of the meta-learning experiments without weighted classes}
  \label{fig2-noweight}
\end{figure}

Figure~\ref{fig2-noweight} shows the results obtained by the ResNet model pre-trained with ImageNet without using the class and distribution balancer algorithm.

The results obtained with these experiments demonstrate the benefit of using knowledge of an unrelated domain to predict the target class of the medical data as compared with the performance obtained with random initialization. This increase in the performance of classification is due to the ability of the model to extract properties that are good to generalize in the medical image dataset.

\begin{figure}
  \centering
  \fbox{
  \includegraphics[scale=0.5]{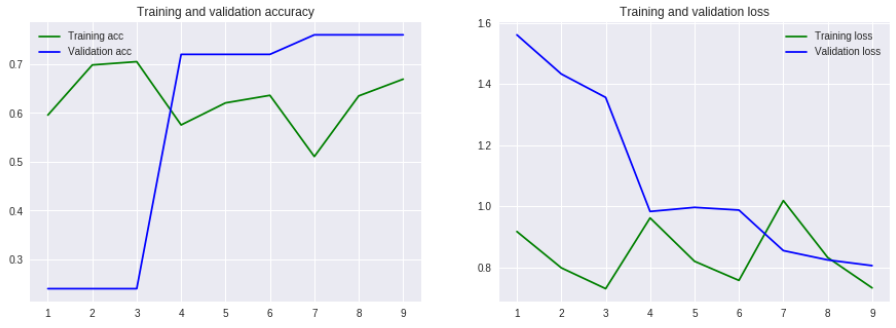}
  }
  \caption{Left image: training and validation loss of the meta-learning experiments using the Class balancer algorithm. Right image: training and validation accuracy.}
  \label{fig3-weight}
\end{figure}

Figure~\ref{fig3-weight} shows the results obtained by using the Class balancer algorithm with the ResNet model pre-trained with ImageNet. This results shows a boost in the performance of the model in \textbf{5 points} in Jaccard similarity index. This increase corresponds to the improvement in the accuracy of the model in the recognition of melanoma cases. An improvement in the validation accuracy and a considerable reduction in the validation loss can be seen as compared with the results obtained without using the Class and distribution balancer algorithm. The summary of the results from these experiments is given in Table~\ref{table4-results}.

\begin{table}
  \caption{Summarized results}
  \label{table4-results}
  \centering
  \begin{tabular}{lll}
  \cmidrule(r){1-3}
    & Jaccard Similarity index  & F-1 score
    \\
    \midrule
    \midrule
    Benchmark & 23.98 & 0.39    
    \\
    Fine-tunning with no weighted classes & 43.03 & 0.54
    \\
    Fine-tunning with weighting class algorithm & \textbf{47.22} & 0.53
    \\
    \bottomrule
    \\
  \end{tabular}
\end{table}

\section{Conclusions}

The results of the benchmarking experiments show that the model performs very well at training, but underperform in the validation set. This behaviour can be explained by the overfitting of the model towards the majority class, since the combined dataset is relatively small.

From the meta-learning experiments using ResNet for pre-training, a general improvement in the accuracy and a reduction in the loss is seen, as compared to the benchmarking experiments with random weights initialization. The results of the meta-learning experiments show that the model improved its generalization on the target dataset, and increased the recognition of the melanoma class.

The results obtained using the class weights generated by the Class and distribution balancer algorithm are interesting, as they show the same performance during training as compared with the experiments with no class weights. However, an increase in the accuracy during validation, and a significant drop in the validation loss was obtained. This boost in performance suggests that the model is sensitive to the covariate shift, therefore this should be considered in the development of meta-learning models for medical domains.

The results from this study suggest that popular deep learning models, such as ResNet, can extract knowledge of data from everyday objects and generalize for the classification of medical data, specifically the skin cancer moles, tackling one of the main challenges in the application of deep learning in medical domains, the scarcity of the data.

\small

\bibliographystyle{apalike}

\end{document}